\documentclass[sigconf]{acmart}
\usepackage{subfigure}
\usepackage{algorithm}
\usepackage{algorithmic}
\usepackage{amsmath}
\usepackage{amsthm}
\usepackage{multirow}
\AtBeginDocument{%
  \providecommand\BibTeX{{%
    \normalfont B\kern-0.5em{\scshape i\kern-0.25em b}\kern-0.8em\TeX}}}

\copyrightyear{2024}
\acmYear{2024}
\setcopyright{acmlicensed}
\acmConference[KDD '24] {Proceedings of the 30th ACM SIGKDD Conference on Knowledge Discovery and Data Mining }{August 25--29, 2024}{Barcelona, Spain.}
\acmBooktitle{Proceedings of the 30th ACM SIGKDD Conference on Knowledge Discovery and Data Mining (KDD '24), August 25--29, 2024, Barcelona, Spain}
\acmISBN{979-8-4007-0490-1/24/08}
\acmDOI{10.1145/3637528.3671922}

\begin{document}

\title{STEMO: Early Spatio-temporal Forecasting with Multi-Objective Reinforcement Learning}

\author{Wei Shao}
\authornote{authors contributed equally to this research.}
\orcid{0000-0002-9873-8331}
\affiliation{%
  \institution{Data61, CSIRO}
  \streetaddress{Research Way}
  \city{Clayton, Victoria}
  \country{Australia}}
\email{phdweishao@gmail.com}

\author{Yufan Kang}
\authornotemark[1]
\affiliation{%
  \institution{RMIT University}
  \city{Melbourne}
  \state{Victoria}
  \country{Australia}
  \postcode{3000}
}
\email{yufan.kang@student.rmit.edu.au}

\author{Ziyan Peng}
\affiliation{%
  \institution{Xidian University}
  \streetaddress{1 Th{\o}rv{\"a}ld Circle}
  \city{Xi'an}
  \country{China}}

\author{Xiao Xiao}
\authornote{Corresponding Author}
\affiliation{%
  \institution{Xidian University}
  \city{Xi'an}
  \country{China}
}
\email{xiaoxiao@xidian.edu.cn
}

\author{Lei Wang}
\affiliation{%
  \institution{Zhejiang University}
  \city{Hangzhou}
  \country{China}}

\author{Yuhui Yang}
\affiliation{%
  \institution{Xidian University}
  \city{Xi'an}
  \country{China}
}

\author{Flora D. Salim}
\affiliation{%
  \institution{University of New South Wales}
  \city{Sydney}
  \country{Australia}}

\renewcommand{\shortauthors}{Shao and Kang, et al.}

\begin{abstract}
Accuracy and timeliness are indeed often conflicting goals in prediction tasks. Premature predictions may yield a higher rate of false alarms, whereas delaying predictions to gather more information can render them too late to be useful. In applications such as wildfires, crimes, and traffic jams, timely forecasting are vital for safeguarding human life and property. Consequently, finding a balance between accuracy and timeliness is crucial. In this paper, we propose an early spatio-temporal forecasting model based on Multi-Objective reinforcement learning that can either implement an optimal policy given a preference or infer the preference based on a small number of samples. The model addresses two primary challenges: 1) enhancing the accuracy of early forecasting and 2) providing the optimal policy for determining the most suitable prediction time for each area. Our method demonstrates superior performance on three large-scale real-world datasets, surpassing existing methods in early spatio-temporal forecasting tasks.
\end{abstract}




\maketitle

\section{Introduction}
Spatio-temporal prediction, an innovative intersection of geographic information systems, statistics, and data science, plays a pivotal role in fields where spatial distribution and temporal progression dictate outcomes. This predictive approach finds wide-ranging applications in domains such as meteorology, epidemiology, traffic control, parking, and urban planning~\cite{yang2021predicting,joseph2019spatiotemporal,li2021dynamic,xia2021spatial, shao2024transferrable, shao2022long}.


\begin{figure}[!ht]
    \centering
    \includegraphics[width=\linewidth]{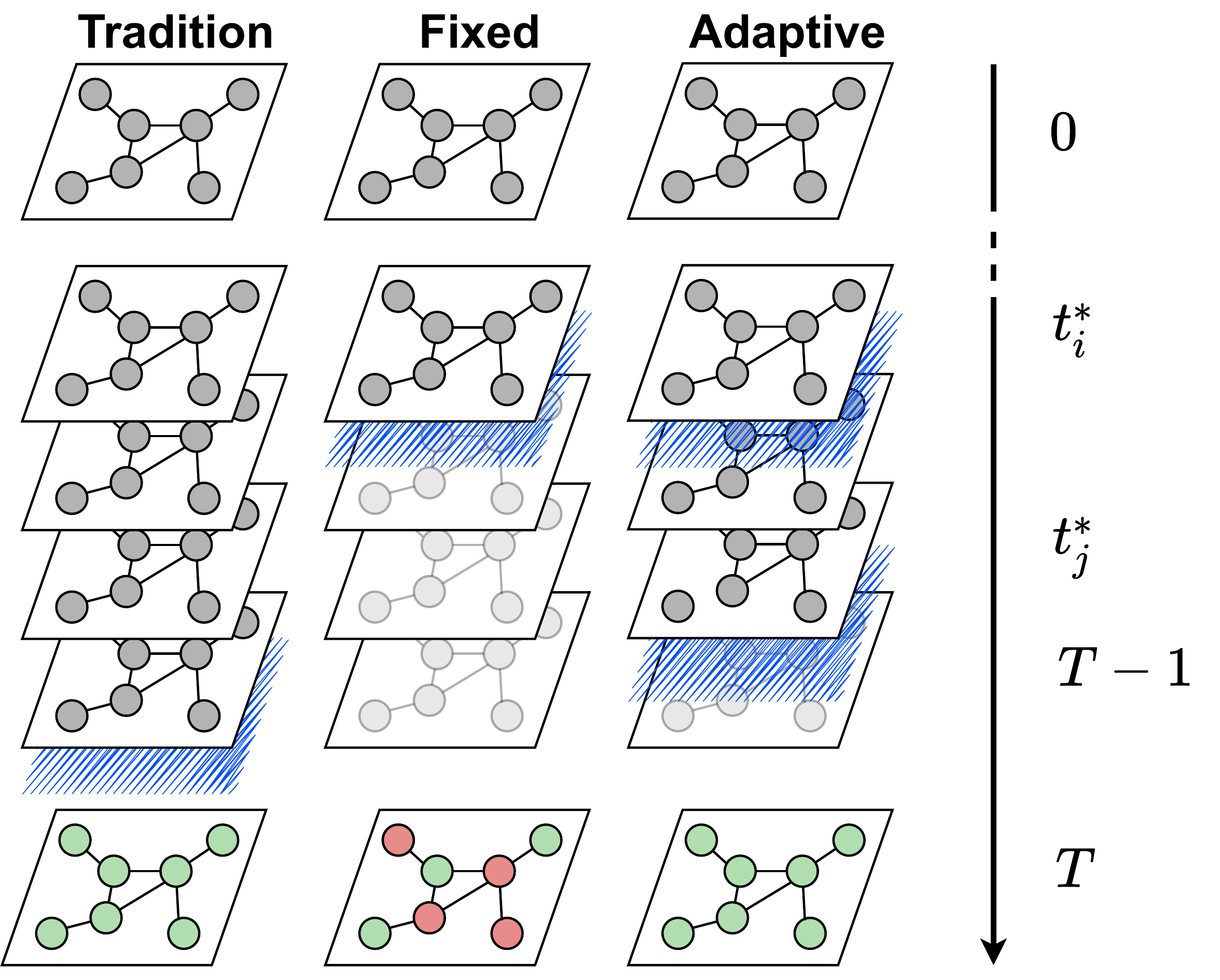}
    \caption{Example of three methods for early spatio-temporal forecasting. Each circle is a node (e.g. sensor) with recorded value (such as speed) over time. The blue plane represents the prediction time, and the recorded values after the blue plane are not used for forecasting. The adaptive early forecasting method adjusts data usage and dynamically determines the prediction time for different nodes. The node colour at time $T$ indicates the accuracy, green indicates that the predicted value matches the ground truth, and red indicates otherwise.}
    \label{introduction}
\end{figure}

Early spatio-temporal forecasting focuses on anticipating future events by examining patterns that change over space and time as early as possible, which is essential to many real-world applications such as epidemiology, environmental studies, and public safety, where early detection can lead to better management or prevention outcomes. For instance, within the context of forecasting the spread of diseases, the timeliness forecasting is considerably more critical than their accuracy. Predictions that are accurate but delayed can lead to the unnecessary loss of thousands of lives. On the other hand, a prediction that is approximately accurate but made in a timely manner can significantly reduce the impact, saving lives.

Figure~\ref{introduction} shows existing approaches to forecast spatio-temporal events are categorised into three distinct strategies: Traditional, Fixed, and Adaptive. For example, Hochreiter \emph{et al.}~\cite{hochreiter1997long}  propose a traditional method that considers the entire recorded data before the forecasting. However, this approach can be time-consuming, particularly with large datasets, and its requirement for complete prior data collection often results in delays, which is unsuitable for time-sensitive applications. Conversely, Li's approach operates within a set observation period~\cite{li2017diffusion}, facilitating fixed timely forecasts within a specific window. Although this strategy enhances promptness, its fixed timing can result in inaccuracies if the observation window is not optimally selected. Moreover, its rigidity in adjusting to temporal data variations and the risk of selection bias in determining the observation window could compromise its precision. The third approach, named adaptive early forecasting, dynamically adjusts the forecasting time based on data characteristics. This idea is inspired by Hartvigsen, who applied this idea in the time-series prediction area~\cite{hartvigsen2019adaptive}. This innovative method aims to strike an optimal balance between forecasting accuracy and timeliness, thereby attempting to optimise both the timeliness and accuracy. Its dynamic adaptability provides a significant edge in handling changes over time.


Although Hartivigsen has tried to apply the concept of early prediction in time-series area~\cite{hartvigsen2019adaptive} and achieved some good results in the medical diagnosis, many challenges remain, especially in the spatio-temporal forecasting. 
(1) \textbf{Real-time Bi-objective Balance in Highly Complex Environment:} Spatio-temporal data demand dynamic solutions, being more complex and subject to rapid changes than time-series data. Traditional multi-objective optimisation methods, which are static and computationally intensive, fail to adapt efficiently across different environments. At the same time, early forecasting necessitates immediate results, as even a one-minute delay can lead to unfavourable outcomes.
(2) \textbf{Spatio-temporal Comprehensive Dependency:} Relying solely on distance correlation for spatio-temporal forecasting may fall short in quickly capturing essential data characteristics. Although distance correlation effectively reflects spatial connections among nodes, it overlooks the crucial temporal aspects embedded within the data, which are vital for precise forecasting. (3) \textbf{Hidden Preferences Discovery:} The equilibrium between timeliness and accuracy varies across tasks, revealing diverse preferences. Identifying these subtle preferences is a complex challenge that demands a profound comprehension of the goals at hand and formulating an ideal balance between promptness and precision for each scenario. This highlights the necessity for refined and flexible strategies in early spatio-temporal forecasting.

We address the aforementioned challenges by introducing a early spatio-temporal forecasting model based on multi-objective reinforcement learning (STEMO). The key contributions of our study can be summarised as follows:
\begin{itemize}
\item We present a multi-objective reinforcement learning framework designed to optimise both the timeliness and accuracy of spatio-temporal forecasting. This approach enhances adaptability to changing patterns through interactive learning from real-time feedback.
\item We introduce a multiple-step similarity matrix, enabling each node to capture the trends from other upstream nodes so that we can estimate the changes of such nodes earlier.
\item We develop a node embedding technique based on biased random walks, increasing the likelihood of visiting nodes with higher similarity and reaching the optimal time. To address the issue of non-uniform object scales or units, we devise a method to discover hidden preferences and employ the entropy weight method.
\end{itemize}

\section{Related Work}
\subsection{Spatio-temporal Prediction}
It is evident that spatio-temporal prediction is of paramount importance in numerous real-world applications~\cite{9204396}, such as weather prediction~\cite{castro2021stconvs2s}, traffic flow prediction~\cite{ermagun2018spatiotemporal}, and earthquake prediction \cite{zhu2021novel}. Traditional time-series methods, including ARIMA \cite{ahmed1979analysis}, have been extensively employed for prediction. With the evolution of machine learning, deep learning methods such as Long Short Term Memory (LSTM) \cite{hochreiter1997long} and Gate Recurrent Unit (GRU) \cite{cho2014properties} have demonstrated superior capabilities in capturing temporal correlations. Convolutional Neural Networks (CNNs)~\cite{lecun1989backpropagation} are frequently utilised to capture spatial correlations in Euclidean space, whereas Graph Convolution Networks (GCNs)~\cite{kipf2016semi} serve to model non-Euclidean relationships among nodes. Despite these advancements, fixed observation windows limit the ability of these models to adjust their predictions in response to evolving data patterns, which could compromise their timeliness and accuracy, especially when dealing with rapidly changing data sources like traffic flow.

\subsection{Early Prediction}
Early prediction methods are techniques employed to make predictions based on specific identifiable characteristics or temporal patterns within a dataset. These methods can be classified into two categories: shapelet-based methods and predictor-based methods. Shapelet-based methods \cite{he2015early,ghalwash2014utilizing} involve finding small sub-parts of the time series data, known as shapelets~\cite{ye2009time}, that can be utilised for prediction. Predictor-based methods \cite{ghalwash2012early,mori2017reliable,mori2017early, shao2023early} involve combining a set of predictors built at different points in time with one or more conditions or trigger functions to evaluate the reliability of predictions and help determine whether they should be considered or discarded. Most of these methods prioritise accuracy over timeliness, inadvertently downplaying the role of prompt predictions. Some methods \cite{mori2017reliable,mori2017early,tavenard2016cost} allow for adjusting the trade-off between accuracy and timeliness with parameters, but these parameters can be difficult to adjust in advance and may require executing the algorithm multiple times to obtain solutions with different trade-offs. In this paper, we propose a Spatio-Temporal Early Prediction model that seeks to improve both the timeliness and accuracy of spatio-temporal predictions, addressing some of the limitations inherent in existing techniques.

\section{Problem Definition}
Consider a graph $\mathcal{G}=\{\mathbf{V},\mathbf{E}\}$, where $\mathbf{V}=\{v_i\}_{i=1}^n$ denotes the set of nodes and $\mathbf{E}=\{e_{ij}\}_{i,j=1}^n$ represents the set of edges. The adjacency matrix $\mathbf{A}\in\mathbb{R}^{n\times n}$  captures the relationships (e.g., distance) between nodes. Recorded values $\mathbf{X}_t=\{x_t^i\}_{i=1}^n$ refer to the measurements obtained from the nodes at time $t$. forecasted values $\widehat{\mathbf{X}}_T=\{\hat{x}_T^i\}_{i=1}^n$ are the estimates of the values at time $T$ based on the graph neural network model. For instance, in a traffic speed forecasting scenario, $x_t^i$ could be the speed at sensor $v_i$ at time $t$, and $\hat{x}_T^i$ could be the forecasted speed at time $T$. The optimal time series $\mathbf{t^*}=\{t^*_i\}_{i=1}^n$ can guide the data observation process. For each node $v_i$, the corresponding optimal time $t_i^*$ is the most appropriate that the forecasting accuracy is maximised and the time cost minimises. We aim to find the optimal time series that balances forecasting accuracy and time cost. The optimal time for node $v_i$ is computed using the following expression:

\begin{equation}
\begin{aligned}
&t^*_i=\mathop{\arg\max}\limits_{t}\left(\log P\left(\hat{x}_T^i|\mathbf{X}_{0:t},\mathcal{G}\right)-\mathrm{Cost}(t)\right). \\
&\text{s.t. }t\in[0,T-1]
\end{aligned}
\end{equation}
The objective function $\log P(\hat{x}_T^i|\mathbf{X}_{0:t},\mathcal{G})$ denotes the log-likelihood of the forecasted value given the past measurements and the graph structure, which can be seen as a measure of the accuracy. The function $\mathrm{Cost}(t)$ represents the time cost of collecting records up to time $t$, and it is assumed to be monotonically increasing as more data is often more expensive to obtain.

\section{Methodology}
\begin{figure}[!ht]
    \centering
    \includegraphics[scale=0.4]{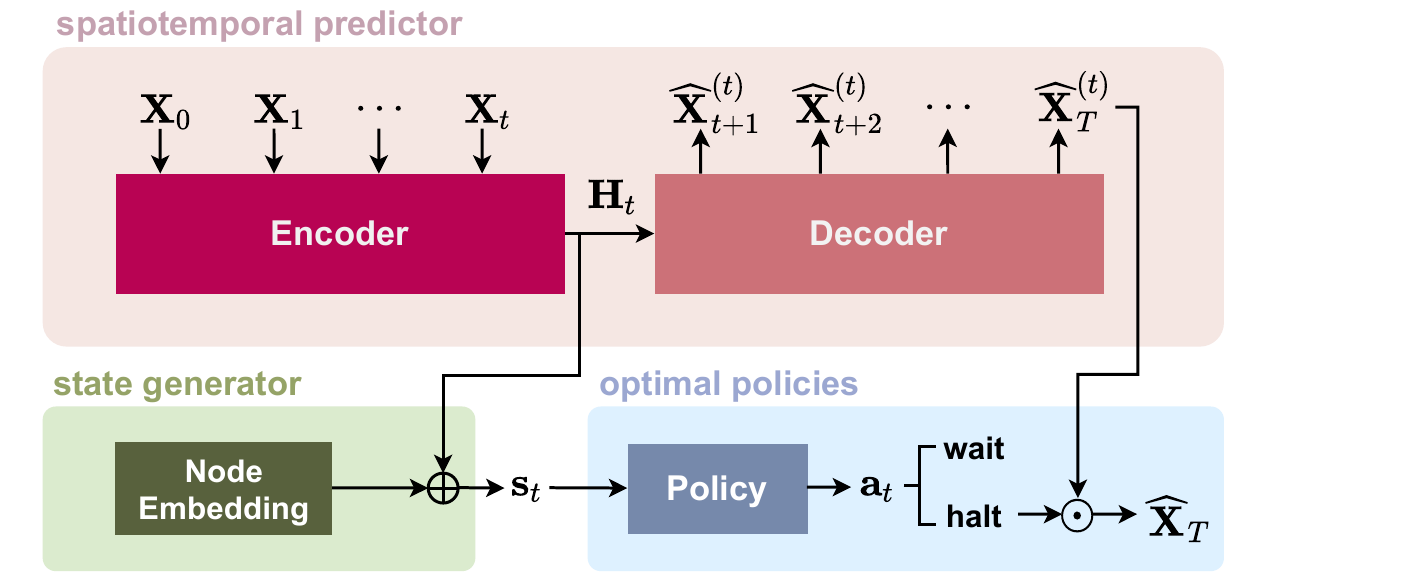}
    \caption{At time $t$, the encoder processes the recorded values $\mathbf{X}_{0:t}$ to extract spatio-temporal features and generate the hidden state $\mathbf{H}_t$. Using $\mathbf{H}_t$, the decoder generates a series of forecasted values, focusing on $\mathbf{\widehat{X}}^{(t)}_{T}$. The state generator concatenates the node embedding result and $\mathbf{H}_t$ to generate the state $\mathbf{s}_t$. The policy utilises $\mathbf{s}_t$ to determine the optimal time for each node $v_i\in\mathbf{V}$ via the action set $\mathbf{a}_t=\{a_t^i\}_{i=1}^n$ (halt or wait). 'Wait' implies that further observation of recorded values is necessary, while 'Halt' implies that time $t$ is the optimal time $t_i^*$ for node $v_i$, and the corresponding forecasted value is recorded in $\mathbf{\widehat{X}}_T$.
    }
    \label{structure}
\end{figure}

As shown in Figure~\ref{structure}, the Early Spatio-Temporal Forecasting model based on Multi-Objective reinforcement learning (STEMO) consists of three main components: a spatio-temporal predictor, a state generator, and optimal policies for determining the optimal time. In addition, we also introduce how to find hidden preferences.

\subsection{The Spatio-temporal Predictor}
The main difference between traditional GCN and the Multi-Graph Convolutional Neural network (MGCN) is the addition of multiple time steps in the similarity matrix calculation in the latter. A standard GCN calculates similarity based on spatial proximity or a single time-point temporal similarity, leading to a limitation in capturing time-evolving trends. MGCN improves this by including multiple time-step similarities, capturing more intricate time-evolving correlations, and facilitating timeliness.

\begin{figure}[!ht]
    \centering
    \subfigure[Time series of nodes $v_i$ and $v_j$. The dotted line represents the series after the optimal time. ]{
\label{Fig.sub.1}
\includegraphics[scale=0.45]{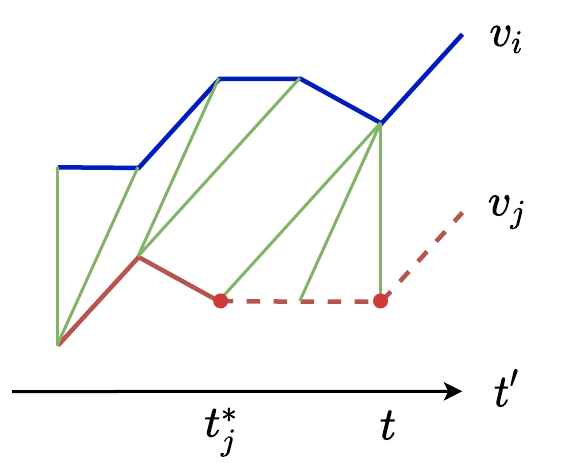}}
\hspace{2mm}
\subfigure[DTW distance between the time series of nodes $v_i$ and $v_j$]{
\label{Fig.sub.2}
\includegraphics[scale=0.45]{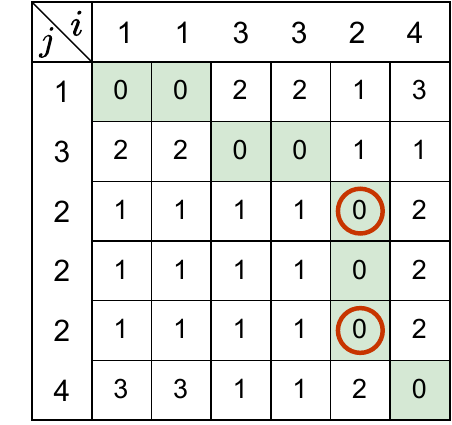}}
\caption{
Figure (a) shows the time series of nodes $v_i$ and $v_j$. The dotted line represents the series after the optimal time, and we only need to observe the solid line to make forecasting. The two solid dots in Figure (a) correspond to the two red circles in Figure (b). Take the red circle below as an example, it corresponds to the DTW distance between $x_{0:t}^i$ and $x_{0:t}^j$, which is calculated along the green grid path. We anticipate that node $i$ will acquire feature $x_{t_j^*}^j$ at time $t$ through MGCN, allowing it to make more precise forecasting earlier.
}

\label{fig:my_label}
\end{figure}

\begin{figure}
    \centering
    \includegraphics[scale=0.35]{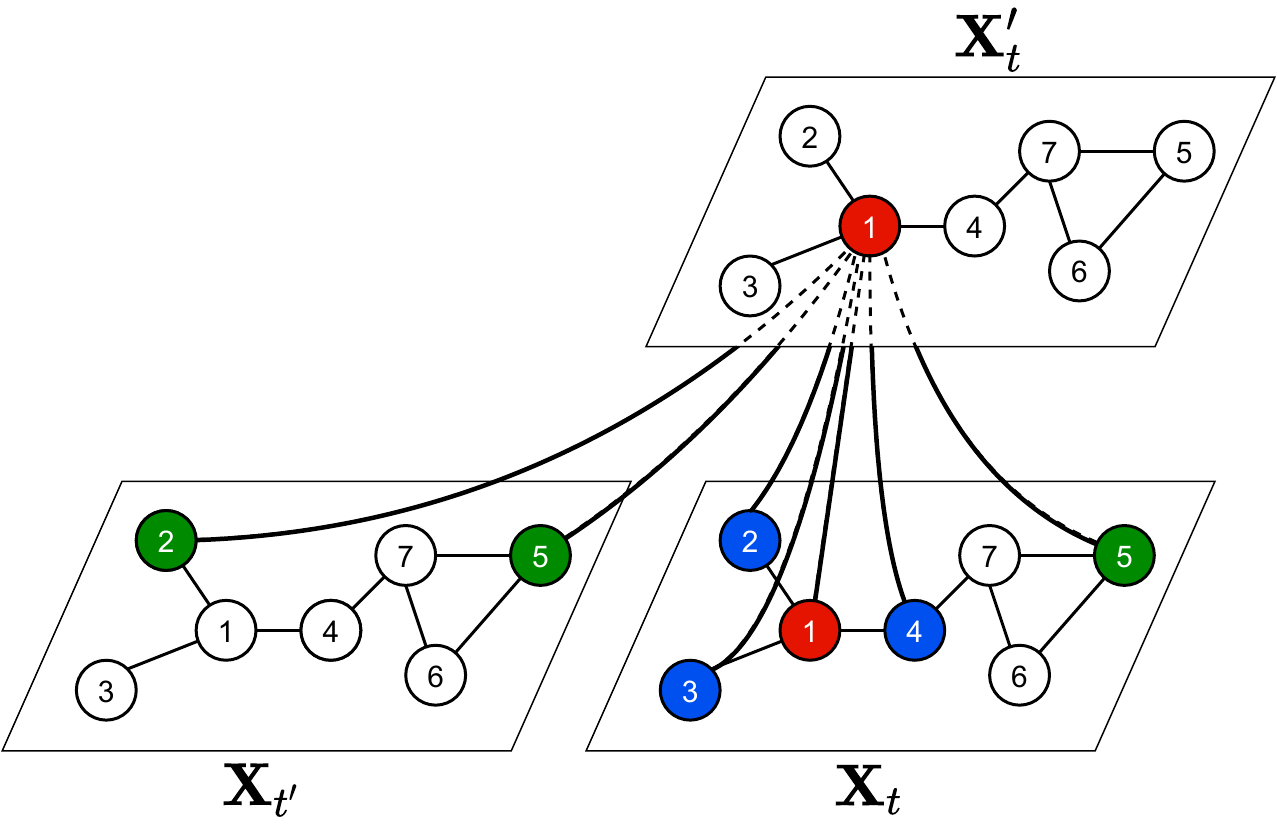}
    \caption{We assume node 1 (red) is the central node. The blue nodes represent spatially close nodes, while the green nodes represent temporally similar nodes. MGCN considers these two kinds of nodes, where spatial and temporal similarities are processed separately but in conjunction.}
    \label{mgcn}
\end{figure}

Specifically, GCN collects equal-length time series of each node, applies the DTW algorithm to each pair of time series data, and uses dynamic time warping (DTW) distance \cite{muller2007dynamic} to create a similarity matrix. DTW is a method that finds an optimal match between two given sequences (e.g., time series) with certain restrictions, which is particularly useful in our setting where we seek to find correlations in temporal data. However, GCN has certain limitations. One limitation is that it does not consider the influence of other nodes before time $t$. As shown in Figure \ref{fig:my_label}, $t$ represents a specific time, and $t'$ is an arbitrary time. If $t'<t$ and $\mathrm{DTW}(x_{0:t'}^j,x_t^i)=0$, this may result from node $v_j$ changing faster than the node $v_i$. In such cases, it would be advantageous for node $v_i$ to incorporate the feature of node $v_j$ at time $t'$, this would enable it to capture the changing trend better and facilitate early forecasting. Moreover, if $t'=t_j^*<t$, this would also allow node $i$ to reach the optimal time more quickly. Another limitation of this approach is that it may produce inaccurate results when $\mathrm{DTW}(x_{t_j^*}^j,x_{t_j^*}^i)>0$, indicating a weak correlation between nodes $v_j$ and $v_i$. This may not always be true and could lead to sub-optimal outcome. The MGCN addresses these limitations, which is designed to optimize both forecasting accuracy and timeliness. As depicted in Figure \ref{mgcn}, the multiple timesteps similarity matrix allows the model to capture correlations better, thus achieving more accurate outcome. In addition, MGCN can incorporate the fast-changing trends from other nodes to make forecasting earlier.

The distance matrix $\mathbf{A}^\mathcal{S}\in\mathbb{R}^{n\times n}$ is calculated according to the geometric distance between nodes:

\begin{equation}
 \mathbf{A}^\mathcal{S}_{i,j}=\\
 \begin{cases}
  \exp\left(-\frac{d_{ij}^2}{\eta^2}\right),&i\neq j\\ 
  0,& i=j
 \end{cases},
\end{equation}
where $d_{ij}$ represents the distance between $v_i$ and $v_j$, and $\eta$ is the parameter that controls the distribution of $\mathbf{A}^\mathcal{S}$.

The multiple timesteps similarity matrix $\mathcal{A}_t^\mathcal{T}\in\mathbb{R}^{(t+1)\times n\times n}$ is computed according to the temporal similarity between nodes:

\begin{equation}
  \mathbf{A}^\mathcal{T}_{t,t',i,j}=
 \begin{cases}
  e^{-\kappa\times\mathrm{DTW}(x_{0:t}^i,x_{0:t'}^j)}&i\neq j\\ 
  0& i=j
 \end{cases},
\end{equation}
where $\kappa$ controls the range of DTW distance and should be set based on the range of relevant conditions.

\begin{equation}
\begin{aligned}
&\mathbf{X}_{t}^{'}=\sigma\left(\sum_{t'=0}^t\mathcal{A}_{t,t'}\mathbf{X}_{t'}\mathcal{W}_{t'}\right)\\
&\mathbf{A}_{t,t'}=
 \begin{cases}
  \mathbf{A}^\mathcal{T}_{t,t'}&t'\leq t\\ 
  (\mathbf{\hat{A}}^\mathcal{S}+\mathbf{A}^\mathcal{T}_{t,t})/2& t'=t
 \end{cases},
\end{aligned}
\end{equation}
where $\mathbf{\hat{A}}^\mathcal{S}=\tilde{\mathbf{D}}^{-\frac{1}{2}}\tilde{\mathbf{A}}^\mathcal{S}\tilde{\mathbf{D}}^{-\frac{1}{2}}$ represents pre-processing step, $\tilde{\mathbf{A}}^{\mathcal{S}}=\mathbf{A}^{\mathcal{S}}+\mathbf{I}$ is the matrix with added self-connections, $\mathbf{I}$ is the identity matrix, $\tilde{\mathbf{D}}=\sum_j\tilde{\mathbf{A}}^{\mathcal{S}}_{i,j}$ is degree matrix, $\sigma(\cdot)$ represents the sigmoid function, $\mathcal{W}_{t'}\in\mathbb{R}^{1\times h}$ is learnable parameters with number of hidden unit $h$.

 The encoder-decoder structure is utilized for its capacity to capture temporal dependencies and handle variable length sequences, as shown in Figure \ref{structure}. In contrast to previous work, the variation in the number of units (consisting of an MGCN and a GRU) allows the model to dynamically adjust its focus to different parts of the time sequence. At time $t$, the encoder has $t+1$ units. The recorded values $\mathbf{X}_t$ are input into MGCN in the unit $t+1$, and the result of graph convolution process $\mathbf{X^{'}}_t$ is then used as input for the GRU. Combining the hidden state $\mathbf{H}_{t-1}$ of GRU output in the previous unit, the GRU outputs hidden state $\mathbf{H}_t$. In practice, in the encoder, only the last prediction unit at each time needs to be calculated. At time $t$, the decoder has $T-t-1$ units, and the last unit outputs the candidate forecasting values $\mathbf{\widehat{X}}_T^{(t)}$.

$\mathbf{\widehat{X}}^{(t)}_T=\mathbf{W}\mathbf{H}_T^{(t)}+\mathbf{b}$ represents candidate forecasted values at time $T$ forecasted at time $t$, where $\mathbf{W}$ and $\mathbf{b}$ are learnable parameters. The spatio-temporal predictor is trained by minimizing the mean absolute error between $\mathbf{\widehat{X}}_T$ and $\mathbf{X}_T$, where forecasted values $\mathbf{\widehat{X}}_T=\sum_{t=0}^{T-1}\mathbf{a}_t\odot\mathbf{\widehat{X}}^{(t)}_T$, and $\odot$ denotes the Hadamard product calculation.

\subsection{State Generator} 
The relationship between nodes is based on temporal similarity and spatial similarity. In essence, if $x^i_{t_i^*}$ and $x^j_{t}$ show similarity, it suggests a possible temporal connection between node $i$ and $j$. We propose to leverage this connection to help node $j$ reach its optimal time more quickly by learning from node $i$. We designed an embedding function $f_t(\cdot)$ for each time $t$, which operates on the features of the node and aims to create a representation of each node at time $t$. 
The state $s^i_t$ captures the current information status of node $i$ at time $t$. It combines the hidden state $\mathbf{H}^i_t\in\mathbb{R}^h$ outputted by the encoder and the node embedding $f_t(v_i)\in\mathbb{R}^e$. The state is used to verify the optimal time by providing comprehensive information for each node at each time.

 The choice of objective function stems from the assumption that a node's state is influenced by its neighbours' states, and the optimisation goal is to maximise the probability of the appearance of the node's neighbours given the node itself:

\begin{equation}
    \max_{f_t}\sum_{{v_i}\in\mathbf{V}}{\log Pr(N(v_i)|f_t(v_i))},
\end{equation}
where $N(v_i)$ represents the set of neighbors of $v_i$ obtained by biased random walk sampling~\cite{Grover2016node2vec:, Nguyen2018BiasedWalk:}.

Biased random walk sampling is used to select a node's neighbours that contribute to its state. The bias term $\alpha_{pq}(t)$ and the transition probability $\min_{t'}\mathcal{A}^\mathcal{T}_{t,t'}$ control the sampling process, favouring nodes that are closer (in terms of reaching their optimal time) and thus more relevant to the early forecasting task. Given the source node $v_i$ and the current node $v_j$, the probability of accessing the next node $v_k$ is represented by the bias term $\alpha_{pq}(t)$ multiplied by the transition probability $ \mathop{\min}\limits_{t'}\mathcal{A}^\mathcal{T}_{t,t'}$, where $\alpha$ represent bias, and is calculated as follows:
\begin{equation}
    \alpha_{pq}(t)=\\
    \begin{cases}
        \frac{1}{p} &\text{if $v_k=v_i$} \\
    \frac{1}{q}&\text{if $t_k^*\leq t$}\\
    1&\text{else}
    \end{cases},
\end{equation}
where $p$ controls the probability of revisiting the traversed node, and $q$ controls the search program to distinguish whether the node reaches the optimal time or not. After the random walk sampling, the remaining steps are the same as in the DeepWalk \cite{perozzi2014deepwalk} algorithm, using the word2vec \cite{church2017word2vec} method to learn the embedding function $f_t(\cdot)$. In essence, nodes are treated as 'words' and node sequences sampled from random walks are treated as 'sentences'. By feeding these sequences into a word2vec model, we can learn an embedding for each node that captures its context within the graph, i.e., its neighbourhood structure.

\subsection{Finding the Optimal Set of Policies}
Since our goal is to induce a single model that can adapt to the entire space of preferences $\Omega$ (a set of different ways decision-makers can prioritise or value the various objectives involved in an optimisation problem), we use a neural network to represent the Q-values $\mathcal{Q}\subseteq(\Omega\rightarrow\mathbb{R}^2)^{\mathcal{S}\times\mathcal{A}}$ with state space $\mathcal{S}$ and action space $\mathcal{A}$. The input to this neural network is a concatenation of the states, actions, and user preferences. At time $t$, the network takes the states $\mathbf{s}_t$ and preferences $\boldsymbol{\omega}\in\Omega$ as input. We estimate the actions $\mathbf{a}_t$ that maximise the output of the network and verify a series of optimal times according to the actions. 

 We use $\varepsilon$-greedy action selection to avoid abundant exploitation as follows: 
\begin{equation}
\label{epsilon}
\mathbf{a}_{t}=\\
 \begin{cases}
  \max_{\mathbf{a}}{\boldsymbol{\omega}^T\mathbf{Q}(\mathbf{s}_t,\mathbf{a},\boldsymbol{\omega};\theta)},&\text{w.p. }1-\varepsilon\\ 
  \text{random},& \text{w.p. }\varepsilon
 \end{cases},
\end{equation}
where $\theta$ indeed refers to the parameters of the Q-network, which are updated during training, action $\mathbf{a}_{t}=\{a_t^i\}_{i=1}^n$ is replaced by a random action with probability $\varepsilon$, and $\varepsilon$ decreases exponentially from 1 to 0 during training. For node $v_i$, $a_{t}^i=1$ indicates that $t$ is the optimal time $t_{i}^*$. If $a_t^i=0,t\in[0,T-1]$, time $T-1$ is the optimal time. 

We designed reward $\mathbf{r}_t=\{r_{t,\text{acc}}^i,r_{t,\text{time}}^i\}_{i=1}^n$ includes accuracy reward and temporal reward. The accuracy reward reflects how well the current prediction and action align with the ground truth, while the temporal reward encourages faster convergence to the optimal time by penalising longer observation time. The specific calculation is as follows:
\begin{equation}
\begin{aligned}
    &r^i_{t,\text{acc}}=-|\widehat{x}_{T}^i-x_{T}^i|,\ r_{t,\text{time}}^i=-\rho t\\
\end{aligned},
\end{equation}
where $\rho$ is the trade-off parameter.

To train the deep neural network, we minimise the following loss function at each step $k$:
\begin{equation}
\begin{aligned}
    &L^A(\theta)=\mathbb{E}_{\mathbf{s}_t,\mathbf{a}_t,\boldsymbol{\omega}}\left[||\mathbf{y}-\mathbf{Q}(\mathbf{s}_t,\mathbf{a}_t,\boldsymbol{\omega};\theta)||_2^2\right]\\   &\mathbf{y}=\mathbf{r}_t+\gamma\arg_Q\max_{\mathbf{a},\boldsymbol{\omega'}}\boldsymbol{\omega}^T\mathbf{Q}(\mathbf{s}_{t+1},\mathbf{a},\boldsymbol{\omega'};\theta_k)
\end{aligned}.
\end{equation}
The $\arg\max$ operation is used to select the action and preference that maximise the future expected total rewards.
However, the optimal frontier contains a large number of discrete policies, which makes the landscape of the loss function non-smooth. To address this problem, we use an auxiliary loss function:
\begin{equation}
L^B(\theta)=\mathbb{E}_{\mathbf{s}_t,\mathbf{a}_t,\boldsymbol{\omega}}\left[|\boldsymbol{\omega}^T\mathbf{y}-\boldsymbol{\omega}^T\mathbf{Q}(\mathbf{s}_t,\mathbf{a}_t,\boldsymbol{\omega};\theta)|\right],
\end{equation}

The final loss function is $L(\theta)=(1-\lambda)L^A(\theta)+\lambda L^B(\theta)$, where $\lambda$ is the weight that controls the trade off $L^A$ and $L^B$.  The value of $\lambda$ is increased gradually from 0 to 1 during training.

\begin{algorithm}[!ht]
\caption{Training of STEMO.}
\textbf{Input}: distance matrix $\mathbf{A}^\mathcal{S}$, a preference sampling distribution $\mathcal{D}$, path $p_\lambda$ for the weight $\lambda$ increasing from 0 to 1. Initialise encoder, decoder, node embedding $f(v_i)$ for $v_i\in\mathbf{V}$, network $\mathbf{Q}$.\\
\textbf{Parameter}: corresponding learnable parameters $\theta_{e}$, $\theta_{d}$, $\theta_{f}$, and $\theta$.\\
\textbf{Output}: learned model
\begin{algorithmic}[1] 
\REPEAT 
\STATE initialised $\mathbf{H}_{-1}$ for encoder. Sample a linear preference $\boldsymbol{\omega}\sim\mathcal{D}$

\FOR{$t = 0,...,T-1$}
\STATE Observe recoded values $\mathbf{X}_t$.\\
\STATE $\mathbf{H}_t=\text{Encoder}(\mathbf{X}_{t}, \mathcal{H}_{t-1}, \mathcal{A}^\mathcal{S}; \theta_{e})$.\\
\STATE $\mathbf{s}_t^i=\mathbf{H}_t||f_t(v_i;\theta_f)\text{, for all } v_i \text{ in } \mathbf{V}$. \\
\STATE $\mathbf{a}_{t}=
 \begin{cases}
  \mathop{\max}\limits_\mathbf{a}{\boldsymbol{\omega}^T\mathbf{Q}(\mathbf{s}_t,\mathbf{a},\boldsymbol{\omega};\theta)},&\text{w.p. }1-\varepsilon\\ 
  \text{random},& \text{w.p. }\varepsilon
 \end{cases}$.\\
\STATE update $\theta_f$ according to equation 5.\\
\STATE initialised go symbol for the decoder.\\
\STATE $\mathbf{\widehat{X}}_{T}^{(t)}=\text{Decoder}(\mathbf{H}_t, \mathcal{A}^\mathcal{S}; \theta_{d})$.\\
 \IF{update network}
 \STATE Sample $N_{\boldsymbol{\omega}}$ preferences $\{\boldsymbol{\omega}_i\sim\mathcal{D}\}$.\\
 \STATE Compute $y_{i}$ for all $1\leq i\leq N_{\boldsymbol{\omega}}$ according to equations 9.\\
 \STATE update parameters $\theta$ by descending its stochastic
gradient $\nabla_\theta L(\theta)$ according to equations 9 and 10:\\
\STATE Increase $\lambda$ along the path $p_\lambda$.\\
 \ENDIF
\ENDFOR
\STATE $\mathbf{\widehat{X}}_T=\sum_{t=0}^{T-1}\mathbf{a}_t\odot\mathbf{\widehat{X}}_{T}^{(t)}$
\STATE update parameters $\theta_{e}$ and $\theta_{d}$ by descending its stochastic gradient according to $\text{MAE }(\mathbf{X}_T,\mathbf{\widehat{X}}_T)$.\\

\UNTIL{stopping criteria is met.}
\end{algorithmic}
\end{algorithm}

\subsection{Finding the Hidden Preference}
Inspired by~\cite{yang2019generalized}, we employ the hidden preference $\boldsymbol{\omega}$, which can greatly influence the policy's decisions.
In the domain of Multi-Objective Reinforcement Learning (MORL), the concept of hidden preferences refers to the underlying, often undisclosed preferences that influence the prioritisation and weighting of multiple competing objectives within a decision-making process. These preferences are deemed 'hidden' as they are not explicitly known a priori and must be inferred from the interactions within the environment. The elucidation of these preferences is pivotal for adapting and optimising policies across diverse tasks where the relative importance of objectives is not directly observable. This necessity arises from the inherent complexity and variability of real-world scenarios, where explicit preference delineation is impractical, thus requiring algorithms capable of discerning and adapting to these latent preferences effectively.

We search the hidden preferences with five steps, (1) The algorithm learns a single parametric policy network that is optimised over the entire space of preferences. This is significant because it allows the system to adapt its policy based on any given preference set, without needing to learn a new policy from scratch. (2) Employing a generalised version of the Bellman equation, which incorporates preferences into the decision-making process. This adaptation enables the system to handle multiple objectives by adjusting the policy according to a set of linear preferences, which can represent the relative importance of each objective. (3) The algorithm uses hindsight experience replay to learn from past decisions. (4) When provided with just scalar rewards and unknown preferences, the algorithm utilises a combination of policy gradient methods and stochastic search over the preference parameters. This approach helps in estimating the hidden preferences that would maximise the expected return, given the observed outcomes.

\section{Experiment}

In this section, we mainly introduce the datasets, baselines and specific parameter settings, etc. Experiments were conducted around comparing the predicted performance of the Early Spatio-Temporal Forecasting model based on Multi-Objective reinforcement learning (STEMO) against baselines, analyzing the effect of each module, and finding the hidden preference.
\subsection{Experiment Settings}
\subsubsection{Datasets}

We conduct experiments on three real-world large-scale datasets: \textbf{METR-LA} \cite{jagadish2014big} The dataset used in this study is a collection of public transportation speed data from 207 sensors on the Los Angeles Expressway. The data was collected using ring detectors, and the sample rate is 5 minutes. The time range for the data is from March 1, 2012 to June 30, 2012. \textbf{EMS}\footnote{\url{https://data.cityofnewyork.us/Public-Safety/EMS-Incident-Dispatch-Data/76xm-jjuj}} 
The emergency dataset used in this study is a collection of data from the New York Fire Department. The dataset contains 145 areas based on postal code and the sample rate is 1 hour. The recorded time range for the data is from January 1, 2011 to November 30, 2011. \textbf{NYPD}\footnote{\url{https://www.kaggle.com/datasets/mrmorj/new-york-city-police-crime-data-historic}} 
The crime dataset used in this study is a collection of data from the New York Police Department. The dataset is divided into 77 regions based on administrative regions. The sample rate is 4 hours, and the recording time range is from January 1, 2014 to December 31, 2015. 
\begin{table}[!ht]
\centering
\caption{Statistics of datasets used in the experiments}
\renewcommand\arraystretch{1.7}
  
  \setlength{\tabcolsep}{5pt}
  \begin{tabular}{p{50pt}p{40pt}p{40pt}p{40pt}}
  \hline
  
  Datasets & Samples &  Nodes & Interval \\ 
  \hline
    METR-LA & 34272 & 207 & 5min \\
    EMS & 7992 & 145 & 1h \\
    NYPD & 4386 & 77 & 4h \\
   
  \hline
\end{tabular}

\label{table1}
\end{table}

Detailed statistics of the datasets are shown in Table \ref{table1}. For NYPD, we take the centroid of the crime site in the area as the location coordinates of the area. For METR-LA, NYPD and EMS, 70\% of the data are used for training, 20\% for testing, and the rest 10\% for verification. 
\begin{table*}[!ht]

\caption{Performance comparison on the three datasets. The average used time percentage is calculated according to $\frac{100\%}{n}\sum_{i=1}^nt^*_i/T$. The optimal time of LSTM, DCRNN, and ASTGCN is a fixed value set in advance, which means that the model is not responsible for determining the optimal time.}
\setlength{\tabcolsep}{1.2mm}
\begin{tabular}{c|c|cc|cc|cc|cc}
\toprule
\multirow{2}{*}{\shortstack{Datasets}}&{\multirow{2}{*}{\shortstack{Average used \\ time percentage}}}&\multicolumn{2}{c}{100\%}&\multicolumn{2}{c}{75\%}&\multicolumn{2}{c}{50\%}&\multicolumn{2}{c}{25\%} \\
 &  & MAE & RMSE  & MAE & RMSE  & MAE & RMSE  & MAE & RMSE  \\
\midrule
\multirow{6}{*}{\rotatebox{90}{METR-LA}}&HA & 7.22 & 9.46  & 7.22 & 9.46  & 7.22 & 9.46  & 7.22 & 9.46 \\
&LSTM & 4.99 $\pm$ 0.07 & 7.48 $\pm$ 0.07  & 7.17$\pm$ 0.11 & 10.21$\pm$ 0.10 & 7.50$\pm$ 0.13  & 10.46 $\pm$ 0.11& 7.72 $\pm$ 0.15 & 10.68 $\pm$ 0.17 \\
&DCRNN & 2.27 $\pm$ 0.06 &\textbf{ 3.94 $\pm$ 0.07}   & 2.91 $\pm$ 0.08 & 5.76 $\pm$ 0.09 & 3.25 $\pm$ 0.10 & 6.68 $\pm$ 0.09 &   3.48 $\pm$ 0.11 & 7.23 $\pm$ 0.10   \\
&ASTGCN & 3.54 $\pm$ 0.16 & 7.39 $\pm$ 0.17 & 5.05 $\pm$ 0.18 & 6.48 $\pm$ 0.18 & 6.65 $\pm$ 0.20 & 12.70 $\pm$ 0.21 & 7.64 $\pm$ 0.23 & 14.40 $\pm$ 0.25   \\
&EARLIEST & 5.01 $\pm$ 0.08 & 7.46 $\pm$ 0.09 & 5.93 $\pm$ 0.10 & 8.21 $\pm$ 0.10 & 6.14 $\pm$ 0.11 & 9.26 $\pm$ 0.10  & 6.85 $\pm$ 0.12 & 10.06 $\pm$ 0.11 \\
&Graph-WaveNet & 2.38 $\pm$ 0.15 & 4.21 $\pm$ 0.08 & 3.22 $\pm$ 0.11 & 6.45 $\pm$ 0.09 & 3.81 $\pm$ 0.07 & 7.70 $\pm$ 0.08 &4.35 $\pm$ 0.16 &8.65 $\pm$ 0.1\\
&\textbf{STEMO} & \textbf{2.23 $\pm$ 0.07}  & 3.96 $\pm$ 0.06 & \textbf{2.37 $\pm$ 0.08} & \textbf{4.45 $\pm$ 0.08}  &3.23 $\pm$ 0.08 & \textbf{5.69 $\pm$ 0.09} & 3.49 $\pm$ 0.09 &  \textbf{6.15 $\pm$ 0.09}\\

\midrule
\multirow{6}{*}{\rotatebox{90}{EMS}}&HA& 1.47 & 2.41  & 1.47 & 2.41 & 1.47 & 2.41  & 1.47 & \textbf{2.41}    \\
&LSTM & 1.61$\pm$ 0.04 & 2.63 $\pm$ 0.05& 1.67$\pm$ 0.04 & 2.78 $\pm$ 0.06 & 1.67 $\pm$ 0.07& 2.79 $\pm$ 0.08 & 1.68 $\pm$ 0.08& 2.78 $\pm$ 0.09   \\
&DCRNN& 1.45$\pm$ 0.03  & 2.60 $\pm$ 0.03 & 1.69 $\pm$ 0.04& 3.19 $\pm$ 0.05 & 1.58 $\pm$ 0.05& 2.92 $\pm$ 0.05 & 1.70 $\pm$ 0.06& 3.16 $\pm$ 0.07  \\
&ASTGCN & 1.49 $\pm$ 0.04 & \textbf{1.95 $\pm$ 0.03}  & 1.78 $\pm$ 0.04 & 2.49 $\pm$ 0.05 & 1.85 $\pm$ 0.05 & 2.84 $\pm$ 0.06 & 2.01 $\pm$ 0.07 & 3.12 $\pm$ 0.08    \\
&EARLIEST & 1.62$\pm$ 0.05 & 2.63 $\pm$ 0.06 & 1.64 $\pm$ 0.05 & 2.67$\pm$ 0.06  & 1.66 $\pm$ 0.06& 2.71 $\pm$ 0.06 & 1.68$\pm$ 0.06 & 2.79$\pm$ 0.07 \\
&Graph-WaveNet & 1.25 $\pm$ 0.07 & 2.42 $\pm$ 0.12 & 1.27 $\pm$ 0.09 & 2.46 $\pm$ 0.16 & 1.29 $\pm$ 0.11 & 2.48 $\pm$ 0.04 & 1.28 $\pm$ 0.06 & 2.49 $\pm$ 0.17 \\
&\textbf{STEMO}  & \textbf{1.29$\pm$ 0.02}  & 2.23 $\pm$ 0.03  & \textbf{1.32$\pm$ 0.03} & \textbf{2.32$\pm$ 0.04} & \textbf{1.40$\pm$ 0.03} & \textbf{2.41$\pm$ 0.03}  & \textbf{1.44$\pm$ 0.03} & 2.49 $\pm$ 0.05\\

\midrule
\multirow{6}{*}{\rotatebox{90}{NYPD}}&HA & 1.54 & 2.00  & 1.54 & 2.00  & 1.54 & \textbf{2.00} & 1.54 & \textbf{2.00}  \\
&LSTM & 1.52 $\pm$ 0.04 & 2.16 $\pm$ 0.04 & 1.61 $\pm$ 0.05& 2.29 $\pm$ 0.06 & 1.73$\pm$ 0.07 & 2.40$\pm$ 0.07 & 1.72$\pm$ 0.08 & 2.40 $\pm$ 0.09 \\
&DCRNN & 1.58 $\pm$ 0.04& 2.16 $\pm$ 0.04& 1.60 $\pm$ 0.05 & 2.21 $\pm$ 0.05 & 1.61 $\pm$ 0.05& 2.22$\pm$ 0.06 & 1.66$\pm$ 0.06 & 2.32$\pm$ 0.07 \\
&ASTGCN & 2.18 $\pm$ 0.06 & 3.08 $\pm$ 0.12 & 2.88 $\pm$ 0.07& 3.31$\pm$ 0.13  & 2.99 $\pm$ 0.08& 3.42$\pm$ 0.13 & 3.69 $\pm$ 0.09& 4.09 $\pm$ 0.14\\
&EARLIEST  & 1.52 $\pm$ 0.05 & 2.17 $\pm$ 0.05& 1.59$\pm$ 0.06 & 2.23$\pm$ 0.07  & 1.64 $\pm$ 0.07& 2.36 $\pm$ 0.07 & 1.65$\pm$ 0.08 & 2.39$\pm$ 0.09\\
&Graph-WaveNet & 1.48 $\pm$ 0.08 & 2.05 $\pm$ 0.11 & 1.5 $\pm$ 0.06 & 2.08 $\pm$ 0.11 & 1.54 $\pm$ 0.15 & 2.14 $\pm$ 0.05 & 1.58 $\pm$ 0.08 & 2.18 $\pm$ 0.02 \\

&\textbf{STEMO} & \textbf{1.36$\pm$ 0.03} & \textbf{1.82$\pm$ 0.05}  & \textbf{1.38$\pm$ 0.04} & \textbf{1.97$\pm$ 0.06} & \textbf{1.45$\pm$ 0.06} & 2.12$\pm$ 0.07 & \textbf{1.49$\pm$ 0.07} & 2.27$\pm$ 0.09 \\
\bottomrule
\end{tabular}

\label{exp1}
\end{table*}%
\begin{table*}[!ht]

\caption{Performance comparison on three datasets (HV and S metrics). $\uparrow$ means that the higher the metric value, the better the model performance, and $\downarrow$ means that the lower the metric value, the better the model performance.}
\setlength{\tabcolsep}{5mm}
\begin{tabular}{c|cc|cc|cc}
\toprule
{\multirow{2}{*}{Datasets}}&\multicolumn{2}{c}{METR-LA}&\multicolumn{2}{c}{EMS}&\multicolumn{2}{c}{NYPD}\\
   & HV$\uparrow$ & S$\downarrow$  &  HV$\uparrow$ & S$\downarrow$  & HV$\uparrow$ & S$\downarrow$  \\
\midrule
HA & 3.71 & -  & 0.56 & -  & \textbf{1.57} & -   \\
LSTM & 2.96 & 1.25  & 0.28 & \textbf{0.34}  & 1.30 & 0.35   \\
DCRNN & 5.88 & 1.33  & 0.12 & 0.55  & 1.38 & 0.34   \\
ASTGCN& 2.41 & 1.77  & 0.26 & 0.59 & 0.36 & 0.53    \\
EARLIEST& 3.91 & 1.08  & 0.33 & \textbf{0.34}  & 1.32 & \textbf{0.33}  \\
Graph-WaveNet& 4.68 & 0.81  & 0.26 & 0.35 & 0.36 & 0.38  \\
\textbf{STEMO}&\textbf{6.73} & \textbf{0.71}  & \textbf{0.58} & \textbf{0.34}  & 1.47 & 0.36  \\

\bottomrule
\end{tabular}

\label{exp3}
\end{table*}%
The proposed model is executed on a Windows system with Nvidia GeForce RTX 3070Ti. \subsubsection{Parameters Settings}
 For our proposed model, we selected parameters based on preliminary experiments and prior research. Specifically, we adopted the Adam optimiser \cite{kingma2014adam} with an initial learning rate of 0.001, which proved to offer a stable and efficient convergence in our preliminary experiments. We set the hidden state dimension $h$ to 12 as it provided a good balance between computational complexity and model performance. Similarly, the batch size is set to 32. The number of preferences $N_\omega=16$, the node embedding dimension $e=4$, the parameter $\kappa=0.005$. In addition, $p=2$, $q=0.5$, $\rho=0.5$, and $T=12$ are all inspired by prior work. The source code is available at \url{https://github.com/coco0106/MO-STEP}.
\subsubsection{Baselines and Metrics}
We compare STEMO with various baselines for spatio-temporal prediction tasks and early prediction tasks, including \textbf{HA}~\cite{liu2004summary},     \textbf{LSTM}~\cite{hochreiter1997long}, \textbf{DCRNN}~\cite{li2017diffusion}, \textbf{ASTGCN}~\cite{guo2019attention}, \textbf{EARLIEST}~\cite{hartvigsen2019adaptive}, \textbf{Graph-WaveNet}~\cite{wu2019graph}.
\subsubsection{Metrics}
We use three popular prediction indicators to evaluate the performance of all models, including mean absolute error (MAE), root mean square error (RMSE), and mean absolute percentage error (MAPE), which are commonly used measures in regression tasks to evaluate the average magnitude of prediction errors.


In addition, we computed the hypervolume (HV) \cite{Zitzler1998AnEA} and the spacing (S) \cite{article1} metrics, which are commonly used to evaluate the performance of multi-objective optimization algorithms. The HV metric measures the volume of the space dominated by the Pareto front found by the algorithm up to a reference point. A higher HV value means that the algorithm has found a set of solutions that dominate a larger part of the target space, which is usually better. S metric measures the distance between adjacent solutions in Pareto frontier. A lower S value is usually better, because it shows that the solutions are more evenly distributed and provide decision makers with a wider range of trade-off options.

\begin{equation}
    \text{HV}=\bigcup_{\phi\in \Phi }V\left(\phi, \varphi\right),
\end{equation}
where $V\left(\phi, \varphi\right)$ represents the hyper volume of the space formed between the solution $\phi$ and the reference point $\varphi$ in the non-dominant solution set $\Phi$, which is the volume of the hypercube constructed with the connecting line between the solution $\phi$ and the reference point $\varphi$ as the diagonal, and the solution with the lowest accuracy and the largest time used percentage in the corresponding data set is used as the reference point.
\begin{equation}
    \text{S}=\sqrt{\frac{1}{|\Phi|-1}\sum_{i=1}^{|\Phi|}(\bar{d}-d_i)^2},
\end{equation}
where $d_i$ represents the minimum distance from the $i^{th}$ solution to other solutions in $\Phi$, and $\bar{d}$ represents the average value of all $d_i$.

\begin{table*}[!ht]

\caption{Some specific elements (the similarity matrix, node embedding, and policy) were excluded from the model and the resulting modified model's performance was compared to that of the complete STEMO model.}
\setlength{\tabcolsep}{1.2mm}
\begin{tabular}{c|ccc|ccc|ccc|ccc}
\toprule
\makebox[0.15\textwidth][c]{\multirow{2}{*}{\shortstack{Average used \\ time percentage}}}&\multicolumn{3}{c}{100\%}&\multicolumn{3}{c}{75\%}&\multicolumn{3}{c}{50\%}&\multicolumn{3}{c}{25\%} \\
 & MAE & RMSE & MAPE & MAE & RMSE & MAPE & MAE & RMSE & MAPE & MAE & RMSE & MAPE \\
\midrule
w/o similarity matrix & 1.38 & 1.98 & 5.21\% & 1.43 & 2.13 & 5.72\% & 1.50 & 2.28 & 6.32\% & 1.57 & 2.43 & 6.65\% \\
w/o node embedding & 1.36 & 1.84 & 4.92\% & 1.40 & 1.96 & 5.67\% & 1.49 & 2.06 & 5.74\% & 1.55 & 2.33 & 6.03\% \\
w/o policy & 1.36 & 1.83 & 4.91\% & 1.39 & 1.98 & 5.72\% & 1.52 & 2.74 & 6.43\% & 1.69 & 2.51 & 6.77\% \\
\midrule
STEMO & \textbf{1.36} & \textbf{1.82} &  \textbf{4.91\%} & \textbf{1.38} & \textbf{1.97} & \textbf{5.01\%} & \textbf{1.45} & \textbf{2.12} & \textbf{5.32\%} & \textbf{1.49} & \textbf{2.27} & \textbf{5.84\%} \\
\midrule
\end{tabular}
\label{exp2}
\end{table*}%

\subsection{Experiment Results}
We evaluate the performance of STEMO against the chosen baseline models in terms of the aforementioned metrics. The impact of each module is quantified through an ablation study, where we sequentially remove one module at a time and measure the change in overall performance. 
In addition, we also reveal hidden preferences. 

\subsubsection{Performance Comparison}
As shown in Table \ref{exp1}, we vary the average used time percentage by adjusting the preference, where the average usage time percentage, defined as the sum of the optimal time instances $t^*_i$ for all $n$ nodes divided by the total time period $T$ and multiplied by 100\%, represents the proportion of optimal time in the total time. This approach allows us to explore different prediction timeliness and assess the corresponding prediction accuracy, where the accuracy is measured by the error between the ground truth and the predicted values obtained by each node at their optimal time. We observe that: \textbf{(1)} When considering the same average used time percentage, the STEMO model exhibits higher accuracy compared to LSTM, DCRNN, ASTGCN, and Graph-WaveNet. The accuracy of the STEMO model shows a lesser decrease when the average usage time percentage experiences an equivalent reduction. This suggests that the STEMO model, which determines the optimal time based on the network's dynamics, can adapt to diverse scenarios, striking a balance between timeliness and accuracy in forecasting. In contrast, the fixed optimal time used by LSTM, DCRNN, ASTGCN, and Graph-WaveNet might not adequately account for the dynamic nature of various scenarios. \textbf{(2)} When the average used time percentage is the same, the STEMO model outperforms EARLIEST in terms of accuracy. This improvement can be attributed to the STEMO model's consideration of spatial and temporal characteristics, especially in scenarios like the METR-LA dataset, where the nodes are densely distributed and exhibit strong correlations. The STEMO model leverages a  Multi-Graph Convolutional Neural network (MGCN) that incorporates multiple time-steps similarity matrices and the distance matrix. This framework effectively captures and utilises the spatio-temporal correlations among the nodes. Additionally, our experiments revealed that ASTGCN exhibits poor performance on the NYPD dataset. This outcome could be attributed to missing values within the dataset and the limited representation ability of the model.

As depicted in Table~\ref{exp3}, above-average HV and lower S metrics for the STEMO model indicate its effectiveness in managing a trade-off between accuracy and timeliness in predictions - our primary objectives in this study.  A higher HV, as observed in the STEMO model, signifies not only better convergence, illustrating the proximity of the solution set to the real Pareto frontier, but also greater extensiveness, indicating a broad coverage of the solution set in the objective space. Additionally, it implies better uniformity, showing an even distribution of individual solutions in the set. This high HV value affirms our assumption that the STEMO model can effectively balance prediction accuracy while maintaining the timeliness of predictions and offer a wide array of diverse solutions. On the other hand, the S metric provides a measure of the dispersion or spread of the solutions in the objective space. A lower S, as displayed by the STEMO model, suggests that the solutions are evenly distributed. This means that the STEMO model offers a variety of optimal times for forecasting, and provides a robust and versatile model for various scenarios. This aligns well with our goal of creating a model that can cater to diverse situations with reliable results across a range of forecasting times.

\subsubsection{Ablation Study}
To assess the contribution of individual components within our proposed STEMO model, we conducted an ablation study using the NYPD dataset, the results of which are presented in Table \ref{exp2}. In the ablation study, we evaluated the following variants of the STEMO model: \textbf{w/o similarity matrix} (without similarity matrix), \textbf{w/o node embedding} (without node embedding), \textbf{w/o policy} (without policy only use fixed value).

As seen from Table \ref{exp2}, the full STEMO model consistently outperforms all variants across different average used time percentages in terms of MAE, RMSE, and MAPE. For instance, when the similarity matrix is removed ('w/o similarity matrix'), the performance degrades, especially at lower average used time percentages. This suggests the importance of the similarity matrix in capturing spatio-temporal information to maintain the model's performance even at lower time percentages. The removal of the node embedding ('w/o node embedding') and the policy network ('w/o policy') also leads to a decrease in performance. This emphasises the significance of the node embedding in capturing the spatial dependencies and the policy network in adaptively determining the optimal time for predictions. Overall, these findings underline the importance of each component and how they collectively contribute to the robust performance of the STEMO model in balancing prediction accuracy and timeliness.

\subsubsection{Revealing hidden preferences}
We modified the parameter $\rho$ to provide vectorised rewards for encoding two different objects: timeliness and accuracy. We used two different tasks (g1, g2) and used only 100 episodes to learn the hidden preferences. The derived hidden preferences are depicted in Table \ref{preference}. The learned preferences of the model are concentrated on the diagonal, indicating that they are in good agreement with the actual potential preferences. For variant g1, the model shows a strong preference (0.67) for timeliness. This result is consistent with the primary goal of g1, which emphasizes making predictions as quickly as possible. g2 primarily focuses on accuracy, as evidenced by the significantly higher preference weight (0.97) for accuracy.
 \begin{table}[!ht]
   \caption{Derived preferences for timeliness and accuracy in the spatio-temporal early prediction model across two task variants (g1 and g2).}
 \setlength{\tabcolsep}{6.5mm}
     \centering
     \begin{tabular}{c|c|c}
     \toprule
         &timeliness & accurancy\\
         \midrule
         g1 & 0.67&0.33\\
         g2& 0.02&0.97\\
     \bottomrule
     \end{tabular}
   
     \label{preference}
 \end{table}

\section{Discussion}
Although the STEMO model shows promise in handling early spatio-temporal forecasting tasks, several areas merit further exploration. First, the model's performance might be restricted if it was primarily validated on a limited set of real-world datasets, which brings into question its generalizability to different scenarios or data types. Future studies could benefit from testing the model on a more diverse datasets from varying domains. Second, the model's complexity might pose challenges in interpretation and computational efficiency. Third, future work should aim to simplify the model or increase its interpretability without sacrificing performance. Additionally, the model's ability to learn hidden preferences requires deeper investigation. 

\section{Conclusion}
We propose the Early Spatio-Temporal Forecasting model based on Multi-Objective reinforcement learning (STEMO) model, specifically designed to address challenges in early spatio-temporal forecasting. The STEMO model leverages the Multi-Graph Convolutional Neural network (MGCN) and Gated Recurrent Unit (GRU) to capture and analyze spatio-temporal correlations. Further, it utilizes the hidden state from the encoder's output in conjunction with node embeddings to adaptively determine the optimal time for forecasting. Additionally, we also explore hidden preferences within our model. Future work will focus on further refining our model, exploring its application in other domains, and addressing any limitations encountered in this study. We believe that our research contributes significantly to the ongoing discourse in the field of spatio-temporal forecasting and will serve as a foundation for future advancements.

\begin{acks}
We acknowledge the support of the Australian Research Council (ARC) Centre of Excellence for Automated Decision-Making and Society (ADM+S) (CE200100005). We also acknowledge the support of National Natural Science Foundation of China (U21A20446). This research/project was undertaken with the assistance of computing resources from RACE (RMIT AWS Cloud Supercomputing) (RMAS00045). Flora Salim would also like to acknowledge edge the support of Cisco’s National Industry Innovation Network (NIIN) Research Chair Program. 
\end{acks}
\bibliographystyle{ACM-Reference-Format}
\bibliography{sample-base}


\begin{thebibliography}{38}


\ifx \showCODEN    \undefined \def \showCODEN     #1{\unskip}     \fi
\ifx \showDOI      \undefined \def \showDOI       #1{#1}\fi
\ifx \showISBNx    \undefined \def \showISBNx     #1{\unskip}     \fi
\ifx \showISBNxiii \undefined \def \showISBNxiii  #1{\unskip}     \fi
\ifx \showISSN     \undefined \def \showISSN      #1{\unskip}     \fi
\ifx \showLCCN     \undefined \def \showLCCN      #1{\unskip}     \fi
\ifx \shownote     \undefined \def \shownote      #1{#1}          \fi
\ifx \showarticletitle \undefined \def \showarticletitle #1{#1}   \fi
\ifx \showURL      \undefined \def \showURL       {\relax}        \fi
\providecommand\bibfield[2]{#2}
\providecommand\bibinfo[2]{#2}
\providecommand\natexlab[1]{#1}
\providecommand\showeprint[2][]{arXiv:#2}

\bibitem[Ahmed and Cook(1979)]%
        {ahmed1979analysis}
\bibfield{author}{\bibinfo{person}{Mohammed~S Ahmed} {and}
  \bibinfo{person}{Allen~R Cook}.} \bibinfo{year}{1979}\natexlab{}.
\newblock \bibinfo{booktitle}{\emph{Analysis of freeway traffic time-series
  data by using Box-Jenkins techniques}}.
\newblock


\bibitem[Castro et~al\mbox{.}(2021)]%
        {castro2021stconvs2s}
\bibfield{author}{\bibinfo{person}{Rafaela Castro}, \bibinfo{person}{Yania~M
  Souto}, \bibinfo{person}{Eduardo Ogasawara}, \bibinfo{person}{Fabio Porto},
  {and} \bibinfo{person}{Eduardo Bezerra}.} \bibinfo{year}{2021}\natexlab{}.
\newblock \showarticletitle{Stconvs2s: Spatiotemporal convolutional sequence to
  sequence network for weather forecasting}.
\newblock \bibinfo{journal}{\emph{Neurocomputing}}  \bibinfo{volume}{426}
  (\bibinfo{year}{2021}), \bibinfo{pages}{285--298}.
\newblock


\bibitem[Cho et~al\mbox{.}(2014)]%
        {cho2014properties}
\bibfield{author}{\bibinfo{person}{Kyunghyun Cho}, \bibinfo{person}{Bart
  Van~Merri{\"e}nboer}, \bibinfo{person}{Dzmitry Bahdanau}, {and}
  \bibinfo{person}{Yoshua Bengio}.} \bibinfo{year}{2014}\natexlab{}.
\newblock \showarticletitle{On the properties of neural machine translation:
  Encoder-decoder approaches}.
\newblock \bibinfo{journal}{\emph{arXiv preprint arXiv:1409.1259}}
  (\bibinfo{year}{2014}).
\newblock


\bibitem[Church(2017)]%
        {church2017word2vec}
\bibfield{author}{\bibinfo{person}{Kenneth~Ward Church}.}
  \bibinfo{year}{2017}\natexlab{}.
\newblock \showarticletitle{Word2Vec}.
\newblock \bibinfo{journal}{\emph{Natural Language Engineering}}
  \bibinfo{volume}{23}, \bibinfo{number}{1} (\bibinfo{year}{2017}),
  \bibinfo{pages}{155--162}.
\newblock


\bibitem[Deb and Jain(2002)]%
        {article1}
\bibfield{author}{\bibinfo{person}{Kalyanmoy Deb} {and} \bibinfo{person}{Sachin
  Jain}.} \bibinfo{year}{2002}\natexlab{}.
\newblock \showarticletitle{Running performance metrics for evolutionary
  multi-objective optimization}.
\newblock  (\bibinfo{date}{07} \bibinfo{year}{2002}).
\newblock


\bibitem[Ermagun and Levinson(2018)]%
        {ermagun2018spatiotemporal}
\bibfield{author}{\bibinfo{person}{Alireza Ermagun} {and}
  \bibinfo{person}{David Levinson}.} \bibinfo{year}{2018}\natexlab{}.
\newblock \showarticletitle{Spatiotemporal traffic forecasting: review and
  proposed directions}.
\newblock \bibinfo{journal}{\emph{Transport Reviews}} \bibinfo{volume}{38},
  \bibinfo{number}{6} (\bibinfo{year}{2018}), \bibinfo{pages}{786--814}.
\newblock


\bibitem[Ghalwash et~al\mbox{.}(2014)]%
        {ghalwash2014utilizing}
\bibfield{author}{\bibinfo{person}{Mohamed~F Ghalwash}, \bibinfo{person}{Vladan
  Radosavljevic}, {and} \bibinfo{person}{Zoran Obradovic}.}
  \bibinfo{year}{2014}\natexlab{}.
\newblock \showarticletitle{Utilizing temporal patterns for estimating
  uncertainty in interpretable early decision making}. In
  \bibinfo{booktitle}{\emph{Proceedings of the 20th ACM SIGKDD international
  conference on Knowledge discovery and data mining}}.
  \bibinfo{pages}{402--411}.
\newblock


\bibitem[Ghalwash et~al\mbox{.}(2012)]%
        {ghalwash2012early}
\bibfield{author}{\bibinfo{person}{Mohamed~F Ghalwash},
  \bibinfo{person}{Du{\v{s}}an Ramljak}, {and} \bibinfo{person}{Zoran
  Obradovi{\'c}}.} \bibinfo{year}{2012}\natexlab{}.
\newblock \showarticletitle{Early classification of multivariate time series
  using a hybrid HMM/SVM model}. In \bibinfo{booktitle}{\emph{2012 IEEE
  International Conference on Bioinformatics and Biomedicine}}. IEEE,
  \bibinfo{pages}{1--6}.
\newblock


\bibitem[Grover and Leskovec(2016)]%
        {Grover2016node2vec:}
\bibfield{author}{\bibinfo{person}{Aditya Grover} {and} \bibinfo{person}{J.
  Leskovec}.} \bibinfo{year}{2016}\natexlab{}.
\newblock \showarticletitle{node2vec: Scalable Feature Learning for Networks}.
\newblock \bibinfo{journal}{\emph{Proceedings of the 22nd ACM SIGKDD
  International Conference on Knowledge Discovery and Data Mining}}
  (\bibinfo{year}{2016}).
\newblock
\urldef\tempurl%
\url{https://doi.org/10.1145/2939672.2939754}
\showDOI{\tempurl}


\bibitem[Guo et~al\mbox{.}(2019)]%
        {guo2019attention}
\bibfield{author}{\bibinfo{person}{Shengnan Guo}, \bibinfo{person}{Youfang
  Lin}, \bibinfo{person}{Ning Feng}, \bibinfo{person}{Chao Song}, {and}
  \bibinfo{person}{Huaiyu Wan}.} \bibinfo{year}{2019}\natexlab{}.
\newblock \showarticletitle{Attention based spatial-temporal graph
  convolutional networks for traffic flow forecasting}. In
  \bibinfo{booktitle}{\emph{Proceedings of the AAAI conference on artificial
  intelligence}}. \bibinfo{publisher}{AAAI Press}, \bibinfo{address}{Honolulu,
  Hawaii, USA}, \bibinfo{pages}{922--929}.
\newblock


\bibitem[Hartvigsen et~al\mbox{.}(2019)]%
        {hartvigsen2019adaptive}
\bibfield{author}{\bibinfo{person}{Thomas Hartvigsen}, \bibinfo{person}{Cansu
  Sen}, \bibinfo{person}{Xiangnan Kong}, {and} \bibinfo{person}{Elke
  Rundensteiner}.} \bibinfo{year}{2019}\natexlab{}.
\newblock \showarticletitle{Adaptive-halting policy network for early
  classification}. In \bibinfo{booktitle}{\emph{Proceedings of the 25th ACM
  SIGKDD International Conference on Knowledge Discovery \& Data Mining}}.
  \bibinfo{pages}{101--110}.
\newblock


\bibitem[He et~al\mbox{.}(2015)]%
        {he2015early}
\bibfield{author}{\bibinfo{person}{Guoliang He}, \bibinfo{person}{Yong Duan},
  \bibinfo{person}{Rong Peng}, \bibinfo{person}{Xiaoyuan Jing},
  \bibinfo{person}{Tieyun Qian}, {and} \bibinfo{person}{Lingling Wang}.}
  \bibinfo{year}{2015}\natexlab{}.
\newblock \showarticletitle{Early classification on multivariate time series}.
\newblock \bibinfo{journal}{\emph{Neurocomputing}}  \bibinfo{volume}{149}
  (\bibinfo{year}{2015}), \bibinfo{pages}{777--787}.
\newblock


\bibitem[Hochreiter and Schmidhuber(1997)]%
        {hochreiter1997long}
\bibfield{author}{\bibinfo{person}{Sepp Hochreiter} {and}
  \bibinfo{person}{J{\"u}rgen Schmidhuber}.} \bibinfo{year}{1997}\natexlab{}.
\newblock \showarticletitle{Long short-term memory}.
\newblock \bibinfo{journal}{\emph{Neural computation}} \bibinfo{volume}{9},
  \bibinfo{number}{8} (\bibinfo{year}{1997}), \bibinfo{pages}{1735--1780}.
\newblock


\bibitem[Jagadish et~al\mbox{.}(2014)]%
        {jagadish2014big}
\bibfield{author}{\bibinfo{person}{Hosagrahar~V Jagadish},
  \bibinfo{person}{Johannes Gehrke}, \bibinfo{person}{Alexandros Labrinidis},
  \bibinfo{person}{Yannis Papakonstantinou}, \bibinfo{person}{Jignesh~M Patel},
  \bibinfo{person}{Raghu Ramakrishnan}, {and} \bibinfo{person}{Cyrus Shahabi}.}
  \bibinfo{year}{2014}\natexlab{}.
\newblock \showarticletitle{Big data and its technical challenges}.
\newblock \bibinfo{journal}{\emph{Commun. ACM}} \bibinfo{volume}{57},
  \bibinfo{number}{7} (\bibinfo{year}{2014}), \bibinfo{pages}{86--94}.
\newblock


\bibitem[Joseph et~al\mbox{.}(2019)]%
        {joseph2019spatiotemporal}
\bibfield{author}{\bibinfo{person}{Maxwell~B Joseph},
  \bibinfo{person}{Matthew~W Rossi}, \bibinfo{person}{Nathan~P Mietkiewicz},
  \bibinfo{person}{Adam~L Mahood}, \bibinfo{person}{Megan~E Cattau},
  \bibinfo{person}{Lise~Ann St.~Denis}, \bibinfo{person}{R~Chelsea Nagy},
  \bibinfo{person}{Virginia Iglesias}, \bibinfo{person}{John~T Abatzoglou},
  {and} \bibinfo{person}{Jennifer~K Balch}.} \bibinfo{year}{2019}\natexlab{}.
\newblock \showarticletitle{Spatiotemporal prediction of wildfire size extremes
  with Bayesian finite sample maxima}.
\newblock \bibinfo{journal}{\emph{Ecological Applications}}
  \bibinfo{volume}{29}, \bibinfo{number}{6} (\bibinfo{year}{2019}),
  \bibinfo{pages}{e01898}.
\newblock


\bibitem[Kingma and Ba(2014)]%
        {kingma2014adam}
\bibfield{author}{\bibinfo{person}{Diederik~P Kingma} {and}
  \bibinfo{person}{Jimmy Ba}.} \bibinfo{year}{2014}\natexlab{}.
\newblock \showarticletitle{Adam: A method for stochastic optimization}.
\newblock \bibinfo{journal}{\emph{arXiv preprint arXiv:1412.6980}}
  (\bibinfo{year}{2014}).
\newblock


\bibitem[Kipf and Welling(2016)]%
        {kipf2016semi}
\bibfield{author}{\bibinfo{person}{Thomas~N Kipf} {and} \bibinfo{person}{Max
  Welling}.} \bibinfo{year}{2016}\natexlab{}.
\newblock \showarticletitle{Semi-supervised classification with graph
  convolutional networks}.
\newblock \bibinfo{journal}{\emph{arXiv preprint arXiv:1609.02907}}
  \bibinfo{volume}{54}, \bibinfo{number}{4} (\bibinfo{year}{2016}),
  \bibinfo{pages}{2645–2656}.
\newblock


\bibitem[LeCun et~al\mbox{.}(1989)]%
        {lecun1989backpropagation}
\bibfield{author}{\bibinfo{person}{Yann LeCun}, \bibinfo{person}{Bernhard
  Boser}, \bibinfo{person}{John~S Denker}, \bibinfo{person}{Donnie Henderson},
  \bibinfo{person}{Richard~E Howard}, \bibinfo{person}{Wayne Hubbard}, {and}
  \bibinfo{person}{Lawrence~D Jackel}.} \bibinfo{year}{1989}\natexlab{}.
\newblock \showarticletitle{Backpropagation applied to handwritten zip code
  recognition}.
\newblock \bibinfo{journal}{\emph{Neural computation}} \bibinfo{volume}{1},
  \bibinfo{number}{4} (\bibinfo{year}{1989}), \bibinfo{pages}{541--551}.
\newblock


\bibitem[Li et~al\mbox{.}(2021)]%
        {li2021dynamic}
\bibfield{author}{\bibinfo{person}{Fuxian Li}, \bibinfo{person}{Jie Feng},
  \bibinfo{person}{Huan Yan}, \bibinfo{person}{Guangyin Jin},
  \bibinfo{person}{Fan Yang}, \bibinfo{person}{Funing Sun},
  \bibinfo{person}{Depeng Jin}, {and} \bibinfo{person}{Yong Li}.}
  \bibinfo{year}{2021}\natexlab{}.
\newblock \showarticletitle{Dynamic graph convolutional recurrent network for
  traffic prediction: Benchmark and solution}.
\newblock \bibinfo{journal}{\emph{ACM Transactions on Knowledge Discovery from
  Data (TKDD)}} (\bibinfo{year}{2021}).
\newblock


\bibitem[Li et~al\mbox{.}(2017)]%
        {li2017diffusion}
\bibfield{author}{\bibinfo{person}{Yaguang Li}, \bibinfo{person}{Rose Yu},
  \bibinfo{person}{Cyrus Shahabi}, {and} \bibinfo{person}{Yan Liu}.}
  \bibinfo{year}{2017}\natexlab{}.
\newblock \showarticletitle{Diffusion convolutional recurrent neural network:
  Data-driven traffic forecasting}.
\newblock \bibinfo{journal}{\emph{arXiv preprint arXiv:1707.01926}}
  (\bibinfo{year}{2017}).
\newblock


\bibitem[Liu and Guan(2004)]%
        {liu2004summary}
\bibfield{author}{\bibinfo{person}{Jing Liu} {and} \bibinfo{person}{Wei Guan}.}
  \bibinfo{year}{2004}\natexlab{}.
\newblock \showarticletitle{A summary of traffic flow forecasting methods [J]}.
\newblock \bibinfo{journal}{\emph{Journal of highway and transportation
  research and development}}  \bibinfo{volume}{3} (\bibinfo{year}{2004}),
  \bibinfo{pages}{82--85}.
\newblock


\bibitem[Mori et~al\mbox{.}(2017a)]%
        {mori2017early}
\bibfield{author}{\bibinfo{person}{Usue Mori}, \bibinfo{person}{Alexander
  Mendiburu}, \bibinfo{person}{Sanjoy Dasgupta}, {and} \bibinfo{person}{Jose~A
  Lozano}.} \bibinfo{year}{2017}\natexlab{a}.
\newblock \showarticletitle{Early classification of time series by
  simultaneously optimizing the accuracy and earliness}.
\newblock \bibinfo{journal}{\emph{IEEE transactions on neural networks and
  learning systems}} \bibinfo{volume}{29}, \bibinfo{number}{10}
  (\bibinfo{year}{2017}), \bibinfo{pages}{4569--4578}.
\newblock


\bibitem[Mori et~al\mbox{.}(2017b)]%
        {mori2017reliable}
\bibfield{author}{\bibinfo{person}{Usue Mori}, \bibinfo{person}{Alexander
  Mendiburu}, \bibinfo{person}{Eamonn Keogh}, {and} \bibinfo{person}{Jose~A
  Lozano}.} \bibinfo{year}{2017}\natexlab{b}.
\newblock \showarticletitle{Reliable early classification of time series based
  on discriminating the classes over time}.
\newblock \bibinfo{journal}{\emph{Data mining and knowledge discovery}}
  \bibinfo{volume}{31}, \bibinfo{number}{1} (\bibinfo{year}{2017}),
  \bibinfo{pages}{233--263}.
\newblock


\bibitem[M{\"u}ller(2007)]%
        {muller2007dynamic}
\bibfield{author}{\bibinfo{person}{Meinard M{\"u}ller}.}
  \bibinfo{year}{2007}\natexlab{}.
\newblock \showarticletitle{Dynamic time warping}.
\newblock \bibinfo{journal}{\emph{Information retrieval for music and motion}}
  (\bibinfo{year}{2007}), \bibinfo{pages}{69--84}.
\newblock


\bibitem[Nguyen and Malliaros(2018)]%
        {Nguyen2018BiasedWalk:}
\bibfield{author}{\bibinfo{person}{Duong Nguyen} {and}
  \bibinfo{person}{Fragkiskos~D. Malliaros}.} \bibinfo{year}{2018}\natexlab{}.
\newblock \showarticletitle{BiasedWalk: Biased Sampling for Representation
  Learning on Graphs}.
\newblock \bibinfo{journal}{\emph{2018 IEEE International Conference on Big
  Data (Big Data)}} (\bibinfo{year}{2018}), \bibinfo{pages}{4045--4053}.
\newblock
\urldef\tempurl%
\url{https://doi.org/10.1109/BigData.2018.8621872}
\showDOI{\tempurl}


\bibitem[Perozzi et~al\mbox{.}(2014)]%
        {perozzi2014deepwalk}
\bibfield{author}{\bibinfo{person}{Bryan Perozzi}, \bibinfo{person}{Rami
  Al-Rfou}, {and} \bibinfo{person}{Steven Skiena}.}
  \bibinfo{year}{2014}\natexlab{}.
\newblock \showarticletitle{Deepwalk: Online learning of social
  representations}. In \bibinfo{booktitle}{\emph{Proceedings of the 20th ACM
  SIGKDD international conference on Knowledge discovery and data mining}}.
  \bibinfo{pages}{701--710}.
\newblock


\bibitem[Shao et~al\mbox{.}(2022)]%
        {shao2022long}
\bibfield{author}{\bibinfo{person}{Wei Shao}, \bibinfo{person}{Zhiling Jin},
  \bibinfo{person}{Shuo Wang}, \bibinfo{person}{Yufan Kang},
  \bibinfo{person}{Xiao Xiao}, \bibinfo{person}{Hamid Menouar},
  \bibinfo{person}{Zhaofeng Zhang}, \bibinfo{person}{Junshan Zhang}, {and}
  \bibinfo{person}{Flora Salim}.} \bibinfo{year}{2022}\natexlab{}.
\newblock \showarticletitle{Long-term Spatio-Temporal Forecasting via Dynamic
  Multiple-Graph Attention}. In \bibinfo{booktitle}{\emph{31st International
  Joint Conference on Artificial Intelligence, IJCAI 2022}}. International
  Joint Conferences on Artificial Intelligence, \bibinfo{pages}{2225--2232}.
\newblock


\bibitem[Shao et~al\mbox{.}(2023)]%
        {shao2023early}
\bibfield{author}{\bibinfo{person}{Wei Shao}, \bibinfo{person}{Ziyan Peng},
  \bibinfo{person}{Yufan Kang}, \bibinfo{person}{Xiao Xiao}, {and}
  \bibinfo{person}{Zhiling Jin}.} \bibinfo{year}{2023}\natexlab{}.
\newblock \showarticletitle{Early Spatiotemporal Event Prediction via Adaptive
  Controller and Spatiotemporal Embedding}. In \bibinfo{booktitle}{\emph{2023
  IEEE International Conference on Data Mining (ICDM)}}. IEEE,
  \bibinfo{pages}{1307--1312}.
\newblock


\bibitem[Shao et~al\mbox{.}(2024)]%
        {shao2024transferrable}
\bibfield{author}{\bibinfo{person}{Wei Shao}, \bibinfo{person}{Yu Zhang},
  \bibinfo{person}{Pengfei Xiao}, \bibinfo{person}{Kyle~Kai Qin},
  \bibinfo{person}{Mohammad~Saiedur Rahaman}, \bibinfo{person}{Jeffrey Chan},
  \bibinfo{person}{Bin Guo}, \bibinfo{person}{Andy Song}, {and}
  \bibinfo{person}{Flora~D Salim}.} \bibinfo{year}{2024}\natexlab{}.
\newblock \showarticletitle{Transferrable contextual feature clusters for
  parking occupancy prediction}.
\newblock \bibinfo{journal}{\emph{Pervasive and Mobile Computing}}
  \bibinfo{volume}{97} (\bibinfo{year}{2024}), \bibinfo{pages}{101831}.
\newblock


\bibitem[Tavenard and Malinowski(2016)]%
        {tavenard2016cost}
\bibfield{author}{\bibinfo{person}{Romain Tavenard} {and}
  \bibinfo{person}{Simon Malinowski}.} \bibinfo{year}{2016}\natexlab{}.
\newblock \showarticletitle{Cost-aware early classification of time series}. In
  \bibinfo{booktitle}{\emph{Joint European conference on machine learning and
  knowledge discovery in databases}}. Springer, \bibinfo{pages}{632--647}.
\newblock


\bibitem[Wang et~al\mbox{.}(2022)]%
        {9204396}
\bibfield{author}{\bibinfo{person}{Senzhang Wang}, \bibinfo{person}{Jiannong
  Cao}, {and} \bibinfo{person}{Philip~S. Yu}.} \bibinfo{year}{2022}\natexlab{}.
\newblock \showarticletitle{Deep Learning for Spatio-Temporal Data Mining: A
  Survey}.
\newblock \bibinfo{journal}{\emph{IEEE Transactions on Knowledge and Data
  Engineering}} \bibinfo{volume}{34}, \bibinfo{number}{8}
  (\bibinfo{year}{2022}), \bibinfo{pages}{3681--3700}.
\newblock
\urldef\tempurl%
\url{https://doi.org/10.1109/TKDE.2020.3025580}
\showDOI{\tempurl}


\bibitem[Wu et~al\mbox{.}(2019)]%
        {wu2019graph}
\bibfield{author}{\bibinfo{person}{Zonghan Wu}, \bibinfo{person}{Shirui Pan},
  \bibinfo{person}{Guodong Long}, \bibinfo{person}{Jing Jiang},
  \bibinfo{person}{Xiaojun Chang}, {and} \bibinfo{person}{Chengqi Zhang}.}
  \bibinfo{year}{2019}\natexlab{}.
\newblock \showarticletitle{Graph WaveNet for Deep Spatial-Temporal Graph
  Modeling}. In \bibinfo{booktitle}{\emph{Proceedings of the 28th International
  Joint Conference on Artificial Intelligence}}. AAAI Press,
  \bibinfo{pages}{1907--1913}.
\newblock


\bibitem[Xia et~al\mbox{.}(2021)]%
        {xia2021spatial}
\bibfield{author}{\bibinfo{person}{Lianghao Xia}, \bibinfo{person}{Chao Huang},
  \bibinfo{person}{Yong Xu}, \bibinfo{person}{Peng Dai},
  \bibinfo{person}{Liefeng Bo}, \bibinfo{person}{Xiyue Zhang}, {and}
  \bibinfo{person}{Tianyi Chen}.} \bibinfo{year}{2021}\natexlab{}.
\newblock \showarticletitle{Spatial-Temporal Sequential Hypergraph Network for
  Crime Prediction with Dynamic Multiplex Relation Learning.}. In
  \bibinfo{booktitle}{\emph{IJCAI}}. \bibinfo{publisher}{International Joint
  Conferences on Artificial Intelligence Organization},
  \bibinfo{pages}{1631--1637}.
\newblock


\bibitem[Yang et~al\mbox{.}(2019)]%
        {yang2019generalized}
\bibfield{author}{\bibinfo{person}{Runzhe Yang}, \bibinfo{person}{Xingyuan
  Sun}, {and} \bibinfo{person}{Karthik Narasimhan}.}
  \bibinfo{year}{2019}\natexlab{}.
\newblock \showarticletitle{A generalized algorithm for multi-objective
  reinforcement learning and policy adaptation}.
\newblock \bibinfo{journal}{\emph{Advances in neural information processing
  systems}}  \bibinfo{volume}{32} (\bibinfo{year}{2019}).
\newblock


\bibitem[Yang et~al\mbox{.}(2021)]%
        {yang2021predicting}
\bibfield{author}{\bibinfo{person}{Suwei Yang}, \bibinfo{person}{Massimo
  Lupascu}, {and} \bibinfo{person}{Kuldeep~S Meel}.}
  \bibinfo{year}{2021}\natexlab{}.
\newblock \showarticletitle{Predicting forest fire using remote sensing data
  and machine learning}. In \bibinfo{booktitle}{\emph{Proceedings of the AAAI
  Conference on Artificial Intelligence}}. \bibinfo{publisher}{arXiv},
  \bibinfo{pages}{14983--14990}.
\newblock


\bibitem[Ye and Keogh(2009)]%
        {ye2009time}
\bibfield{author}{\bibinfo{person}{Lexiang Ye} {and} \bibinfo{person}{Eamonn
  Keogh}.} \bibinfo{year}{2009}\natexlab{}.
\newblock \showarticletitle{Time series shapelets: a new primitive for data
  mining}. In \bibinfo{booktitle}{\emph{Proceedings of the 15th ACM SIGKDD
  international conference on Knowledge discovery and data mining}}.
  \bibinfo{pages}{947--956}.
\newblock


\bibitem[Zhu et~al\mbox{.}(2021)]%
        {zhu2021novel}
\bibfield{author}{\bibinfo{person}{Daoye Zhu}, \bibinfo{person}{Yi Yang},
  \bibinfo{person}{Fuhu Ren}, \bibinfo{person}{Shunji Murai},
  \bibinfo{person}{Chengqi Cheng}, {and} \bibinfo{person}{Min Huang}.}
  \bibinfo{year}{2021}\natexlab{}.
\newblock \showarticletitle{Novel Intelligent Spatiotemporal Grid Earthquake
  Early-Warning Model}.
\newblock \bibinfo{journal}{\emph{Remote Sensing}} \bibinfo{volume}{13},
  \bibinfo{number}{17} (\bibinfo{year}{2021}), \bibinfo{pages}{3426}.
\newblock


\bibitem[Zitzler and Thiele(1998)]%
        {Zitzler1998AnEA}
\bibfield{author}{\bibinfo{person}{E. Zitzler} {and} \bibinfo{person}{L.
  Thiele}.} \bibinfo{year}{1998}\natexlab{}.
\newblock \showarticletitle{An evolutionary algorithm for multiobjective
  optimization: the strength Pareto approach}, Vol.~\bibinfo{volume}{43}.
\newblock


\end{thebibliography}









\end{document}